\documentclass[10pt,twocolumn,letterpaper]{article}

\usepackage{iccv}
\usepackage{times}
\usepackage{epsfig}
\usepackage{graphicx}
\usepackage{amsmath}
\usepackage{amssymb}
\usepackage{subfigure}
\usepackage{booktabs}

\usepackage{bbm}
\usepackage{mathrsfs}
\usepackage{enumitem}

\DeclareMathOperator*{\argmin}{arg\,min}

\usepackage[pagebackref=true,breaklinks=true,letterpaper=true,colorlinks,bookmarks=false]{hyperref}

\iccvfinalcopy %

\def\iccvPaperID{5484} %

\usepackage{xcolor}

\definecolor{myRed}{rgb}{1,0.0,0}

\definecolor{myBleu}{rgb}{0.0,0.0,1}

\definecolor{myOrange}{rgb}{1,0.5,0}

\ificcvfinal\fi

\begin{document}

\title{Rethinking Transformer-based Set Prediction for Object Detection}

\author{Zhiqing Sun$^{*}$ \quad Shengcao Cao\thanks{indicates equal contribution.} \quad Yiming Yang \quad  Kris Kitani\\
Carnegie Mellon University\\
{\tt\small \{zhiqings,shengcao,yiming,kkitani\}@cs.cmu.edu}
}

\maketitle
\ificcvfinal\thispagestyle{empty}\fi

\begin{abstract}

DETR is a recently proposed Transformer-based method which views object detection as a set prediction problem and achieves state-of-the-art performance %
but demands extra-long training time to converge.
In this paper, we investigate the causes of the optimization difficulty in the training of DETR.
Our examinations reveal several factors contributing to the slow convergence of DETR, primarily the issues with the Hungarian loss and the Transformer cross-attention mechanism. To overcome these issues we propose two solutions, namely, TSP-FCOS (Transformer-based Set Prediction with FCOS) and TSP-RCNN (Transformer-based Set Prediction with RCNN). 
Experimental results show that the proposed methods not only converge much faster than the original DETR, but also significantly outperform DETR and other baselines in terms of detection accuracy.
Code is released at \url{https://github.com/Edward-Sun/TSP-Detection}.
\end{abstract}

\section{Introduction}

Object detection aims at finding all objects of interest in an image and predicting their category labels and bounding boxes, which is essentially a set prediction problem, as the ordering of the predicted objects is not required.
Most of the state-of-the-art neural detectors
\cite{liu2016ssd, redmon2016you, lin2017focal,redmon2017yolo9000,zhang2020bridging,ren2015faster, he2017mask}
are developed in a detect-and-merge fashion that is, instead of directly optimizing the predicted set in an end-to-end fashion, those
methods usually first make predictions on a set of region proposals or sliding windows, and then perform a post-processing step (e.g., ``non-maximum suppression'' or NMS) for merging the
the detection results in different proposals or windows that might belong to the same object.
As the detection model is trained agnostically with respect to the merging step, the model optimization in those object detectors is not end-to-end and arguably sub-optimal.

DEtection TRansformer (DETR) \cite{carion2020end} is recently proposed as the first fully end-to-end object detector. It uses Transformer \cite{vaswani2017attention} to directly output a final set of predictions without
further post-processing.
However, it takes extra-long training time
to converge. For example, the popular Faster RCNN model \cite{ren2015faster} only requires about 30 epochs to convergence, but DETR needs
500 epochs, which takes at least 10 days on 8 V100 GPUs. Such expensive training cost would be practically prohibitive in large applications.  Therefore, in what manner should we accelerate the training process
towards fast convergence
for DETR-like Transformer-based detectors is a challenging research question and is the main focus of this paper.

For analyzing the causes of DETR's optimization difficulty we conduct extensive experiments and find
that the cross-attention module, by which the Transformer decoder obtains object information from images,
is mainly responsible for the slow convergence. 
In pursuit of faster convergence, we further examine an encoder-only version of DETR by removing the cross-attention module.
We find that the encoder-only DETR yields a substantial improvement for the detection of small objects in particular but sub-optimal performance on large objects.
In addition, our analysis shows that the instability of the bipartite matching in DETR's Hungarian loss also contributes to the slow convergence.

Based on the above analysis we propose two models for significantly accelerating the training process of Transformer-based set prediction methods, both of which can be regarded as improved versions of encoder-only DETR with feature pyramids \cite{lin2017feature}.
Specifically, we present TSP-FCOS (Transformer-based Set Prediction with FCOS) and TSP-RCNN (Transformer-based Set Prediction with RCNN), which are inspired by a classic one-stage detector FCOS \cite{tian2019fcos} (Fully Convolutional One-Stage object detector) and a classic two-stage detector Faster RCNN \cite{ren2015faster}, respectively. A novel Feature of Interest (FoI) selection mechanism is developed in TSP-FCOS to help Transformer encoder handle multi-level features. To resolve the instability of the bipartite matching in the Hungarian loss, we also design a new bipartite matching scheme for each of our two models for accelerating the convergence in training. In our evaluation
on the COCO 2017 detection benchmark \cite{lin2014microsoft} %
the proposed methods not only converge much faster than the original DETR, but also significantly outperform DETR and other baselines in terms of detection accuracy.

\section{Background}
\subsection{One-stage and Two-stage Object Detectors}

 Most modern object detection methods can be divided into two categories: One-stage detectors and two-stage detectors. Typical one-stage detectors \cite{liu2016ssd, redmon2016you, lin2017focal} directly make predictions based on the extracted feature maps and (variable-sized)
 sliding-window locations in a image, while two-stage detectors \cite{ren2015faster, he2017mask} first generate region proposals based on sliding-window locations and then refine the detection for each proposed region afterwards.
 In general, two-stage detectors are more accurate but also computationally more expensive than one-stage detectors. 
 Nevertheless, both kinds of detectors are developed in a detection-and-merge fashion, \ie, they require a post-processing step to ensure that each detected object has only one region instead of multiple overlapping regions as detection results.  In other words, many state-of-the-art object detection methods do not have
 an end-to-end training objective with respect to set prediction.

\subsection{DETR with an End-to-end Objective} \label{sec:set}
\label{sec:set prediction}
Different from the aforementioned popular %
object detectors, DEtection TRansformer (DETR) \cite{carion2020end} presents the first method with an end-to-end optimization objective for set prediction.
Specifically, it formulates the loss function via a bipartite matching mechanism.
Let us denote by $y = \{y_i\}_{u=1}^M$ the ground truth set of objects, and $\hat{y} = \{\hat{y_i}\}_{u=1}^N$ the set of predictions. Generally we have $M < N$, so we pad $y$ to size $N$ with $\emptyset$ (no object) and denote it by $\bar{y}$. The loss function, namely the Hungarian loss, is defined as:
\begin{equation} \label{eq:loss}
    \mathcal{L}_{\text{Hungarian}}(\bar{y}, \hat{y}) = \sum_{i=1}^N \left[
    \mathcal{L}_{class}^{i, \hat{\sigma}(i)} + \mathbbm{1}_{\{\bar{y}_i \not= \emptyset\}} \mathcal{L}_{box}^{i, \hat{\sigma}(i)}
    \right]
\end{equation}
where $\mathcal{L}_{class}^{i, \hat{\sigma}(i)}$ and $\mathcal{L}_{box}^{i, \hat{\sigma}(i)}$ are the classification loss and bounding box regression loss, respectively, 
between the $i^{th}$ ground truth and the $\hat{\sigma}(i)^{th}$ prediction.
And $\hat{\sigma}$ is the optimal bipartite matching between padded ground-truth set $\bar{y}$ and prediction set $\hat{y}$ with lowest matching cost:
\begin{equation} \label{eq:match}
    \hat{\sigma} = \argmin_{\sigma \in \mathfrak{S}_N} \sum_{i=1}^N \mathcal{L}_{\text{match}}(\bar{y}_i, \hat{y}_{\sigma(i)})
\end{equation}
where $\mathfrak{S}_N$ is the set of all $N$ permutations and $\mathcal{L}_{\text{match}}$ is a pair-wise matching cost.

DETR \cite{carion2020end} uses %
an encoder-decoder Transformer \cite{vaswani2017attention} framework %
built upon the CNN backbone.
The Transformer encoder part
processes the flattened deep features\footnote{In this paper, we use ``feature points'' and ``features'' interchangeably.} from the CNN backbone. Then, the non-autoregressive decoder part takes the encoder's outputs and a set of learned object
query vectors as the input, and predicts the category labels and bounding boxes accordingly as the detection output. The cross-attention module plays an important role in the decoder by attending to different locations in the image for different object queries.
We refer the reader unfamiliar with the Transformer concepts to the appendix.
The attention mechanism in DETR eliminates the need for NMS post-processing because the self-attention component can learn to remove duplicated detection, \ie, its Hungarian loss (Equation~\ref{eq:loss}) encourages one target per object in the bipartite matching.

Concurrent to our work, some variants of DETR have been proposed to improve its training efficiency and accuracy. Deformable DETR \cite{zhu2020deformable} proposes to integrate the concept of deformable convolution and attention modules, to implement a sparse attention mechanism on multi-level feature maps. UP-DETR \cite{dai2020up} leverages an unsupervised pre-training task named random query patch detection to improve the performance of DETR when it is fine-tuned on down-stream tasks. Compared to these work, we explore further simplifying the detection head design with encoder-only Transformer.

\subsection{Improving Ground-truth Assignments}

The Hungarian loss in DETR can be viewed as an end-to-end way to assign ground-truth labels to the system predictions.  Prior to DETR, heuristic rules have been tried for this task \cite{girshick2015fast, ren2015faster,redmon2016you}.
There are a few other prior work that try to improve the heuristic ground-truth assignment rules. \cite{zhang2019freeanchor} formulates an MLE procedure to learn the matching between sliding windows and ground truth objects. \cite{rezatofighi2019generalized} proposes a generalized IoU which provides a better metric. %
Nevertheless, those methods do not directly optimize a set-based objective and still require an NMS post-processing step.

\subsection{Attention-based Object Detection}

Attention-based %
modeling has been the current workhorse in the Natural Language Processing (NLP) domain \cite{vaswani2017attention, devlin2018bert}, and is becoming increasingly popular in recent object detection research.
Before the invention of DETR, \cite{hu2018relation} proposes an attention-based module to model the relation between objects, which can be inserted into existing detectors and leads to better recognition and less duplication. \cite{qin2019thundernet} uses a Spatial Attention Module %
to re-weight feature maps for making foreground features standing out.
\cite{chi2020relationnet++} uses a Transformer-like attention-based module to bridge different forms of representations.
But, none of those methods have tried an end-to-end set prediction objective.

\section{What Causes the Slow Convergence of DETR?}

To pin down the main factors we ran a set of experiments with DETR and its variants which are built on top of the ResNet-50 backbone and evaluated on the COCO 2017 validation set.

\begin{figure}
    \centering
    \includegraphics[width=0.7\linewidth]{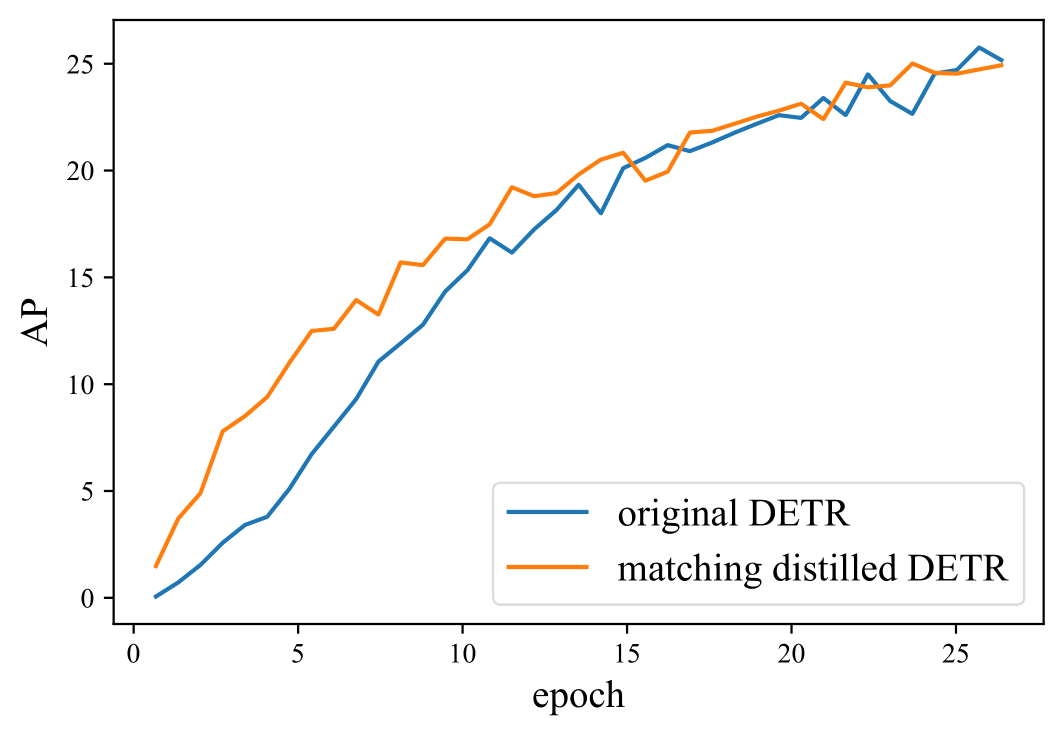}
    \caption{AP results on COCO validation set: original DETR v.s. matching distilled DETR. We can see that matching distillation accelerates the training of DETR in the first few epochs.}
    \label{fig:distill_DETR}
\end{figure}

\subsection{Does Instability of the Bipartite Matching Affect Convergence?} \label{sec:biparitite}

As a unique component in DETR, the Hungarian loss based on the bipartite matching (Section \ref{sec:set prediction}) could be unstable due to the following reasons:
\begin{itemize}[topsep=1.5pt]
     \setlength{\itemsep}{1.5 pt}
     \setlength{\parskip}{1.5 pt}
    \item The initialization of the bipartite matching is essentially random;
    \item The matching instability would be caused by noisy conditions in different training epochs.
\end{itemize}

To examine the effects of these factors, %
we propose a new training strategy for DETR, namely matching distillation. That is, we use a well pre-trained DETR as the teacher model, whose predicted bipartite matching is treated as the ground-truth label assignment for the student model.  All stochastic modules in the teacher model (\ie, dropout \cite{srivastava2014dropout} and batch normalization \cite{ioffe2015batch}) are turned off to ensure the provided matching is deterministic, which eliminates the randomness and instability of the bipartite matching and hence in the Hungarian loss.

We evaluated both the original DETR and matching distilled DETR. Figure~\ref{fig:distill_DETR} shows the results with the first 25 epochs. We can see that the matching distillation strategy does help the convergence of DETR in the first few epochs. %
However, such effect becomes insignificant after around 15 epochs.
This means that the instability in the bipartite matching component of DETR only contributes partially to the slow convergence (especially in the early training stage) but not necessarily the main reason.

\subsection{Are the Attention Modules the Main Cause?}

Another %
distinct part of DETR in comparison with
other modern object detectors is its use of the Transformer modules, where the %
Transformer attention maps are nearly uniform in the initialization stage, but gradually become more and more sparse during the training process towards the convergence.
Prior work \cite{jiang2020convbert} shows that replacing some attention heads in BERT \cite{devlin2018bert} with sparser modules (\eg, convolutions) can significantly accelerate its training.
Therefore, it is natural for us to wonder how much the sparsity dynamics of Transformer attention modules in DETR contribute to its slow convergence.

\begin{figure}
    \centering
    \includegraphics[width=0.9\linewidth]{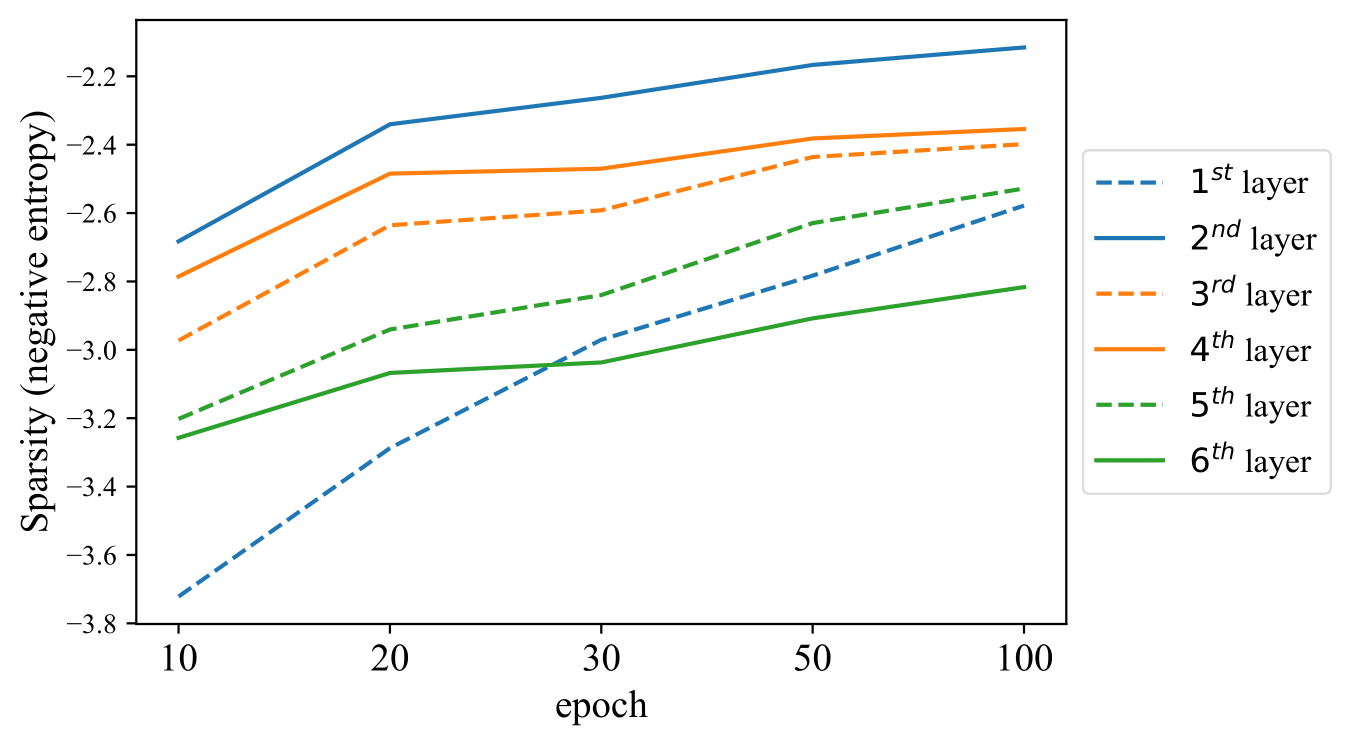}
    \caption{
    Sparsity (negative entropy) of Transformer cross-attention in each layer, obtained by evaluation on the COCO validation data. Different line styles represent  different layers. We can see that the sparsity consistently increases, especially for the $1^{st}$ cross-attention layer between encoder and decoder.
    }
    \label{fig:attention}
\end{figure}

In analyzing the effects of the DETR's attention modules on its optimization convergence, we focus on  
the sparsity dynamics of the cross-attention part in particular, because the cross-attention module is a crucial module where object queries in the decoder obtain object information from the encoder. Imprecise (under-optimized) cross-attention may not allow the decoder to extract accurate context information from images, and thus results in poor localization especially for small objects.

We collect the attention maps of cross-attention when evaluating the DETR model at different training stages. As attention maps can be interpreted as probability distributions, we use negative entropy as an intuitive measure of sparsity. Specifically, given a $n\times m$ attention map $\mathbf{a}$, we first calculate the sparsity of each source position $i \in [n]$ by $\frac{1}{m}\sum_{j=1}^m P(a_{i,j}) \log P(a_{i,j})$, where $a_{i,j}$ represents the attention score from source position $i$ to target position $j$.
Then we average the sparsities for all attention heads and all source positions in each layer. The masked positions \cite{carion2020end} are not considered in the computation of sparsity.

Figure~\ref{fig:attention} shows the sparsities with respect to different epochs at several layers. %
we can see that the sparsity of cross-attention consistently increases and does not reach a plateau even after 100 training epochs.
This means that the cross-attention part of DETR is more dominating a factor for the slow convergence, compared to the early-stage bipartite-matching instability factor we discussed before.

\subsection{Does DETR Really Need Cross-attention?} \label{sec:encoder-only}

Our next question is: Can we remove the cross-attention module from DETR for faster convergence but without sacrificing its prediction power in object detection?  We answer this question by designing
an encoder-only version of DETR and comparing its convergence curves with the original DETR.

In the original DETR, %
the decoder is responsible for producing the detection results (category label and bounding box) per object query.  In contrast, the encoder-only version of DETR (introduced by us) directly uses
the outputs of Transformer encoder for object prediction.
Specifically, for a $H \times W$ image with a $\frac{H}{32} \times \frac{W}{32}$ Transformer encoder feature map, each feature 
is fed into a detection head to predict a detection result.
Since the %
encoder self-attention is essentially identical to the self-attention in a non-autoregressive decoder, a set prediction training is still feasible for encoder-only DETR. More details of encoder-only DETR can be found in the appendix. Figure~\ref{fig:meta} compares the %
original DETR and the encoder-only DETR, and two of our newly proposed models (TSP-FCOS and TSP-RCNN) which are
described in the next section.

\begin{figure}
    \centering
    \includegraphics[width=0.9\linewidth]{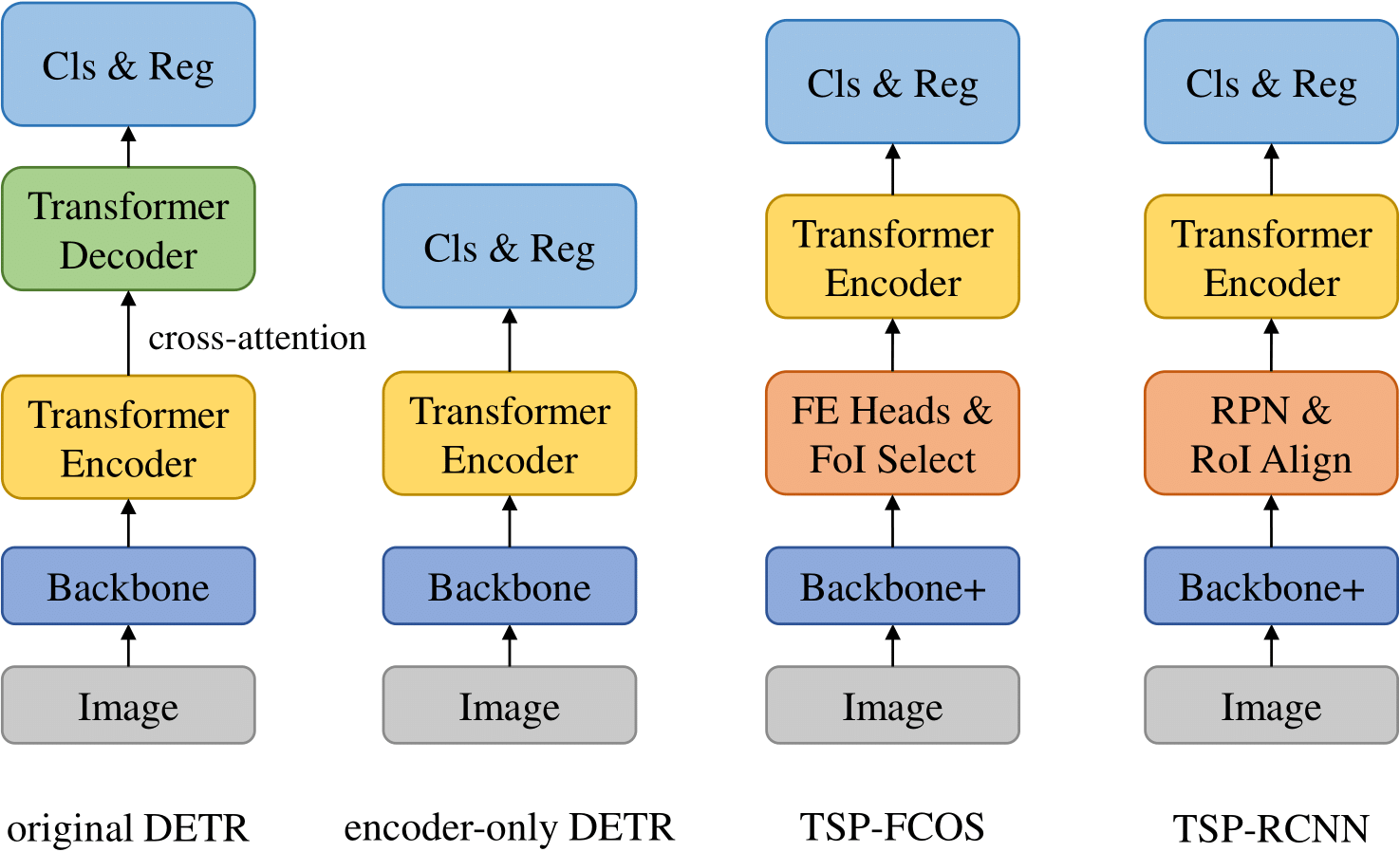}
    \caption{Illustration of original DETR, encoder-only DETR, TSP-FCOS, and TSP-RCNN, where \mbox{Backbone+}, \mbox{FE Heads}, \mbox{RPN}, \mbox{Cls \& Reg} represents ``Backbone + FPN'', ``Feature Extraction Heads (Subnets)'', ``Region Proposal Network'', ``Classification \& Regression'', respectively.
    A more detailed illustration of TSP-FCOS and TSP-RCNN can be found in Figure~\ref{fig:main}.
    }
    \label{fig:meta}
\end{figure}

\begin{figure}
    \centering
    \includegraphics[width=0.9\linewidth]{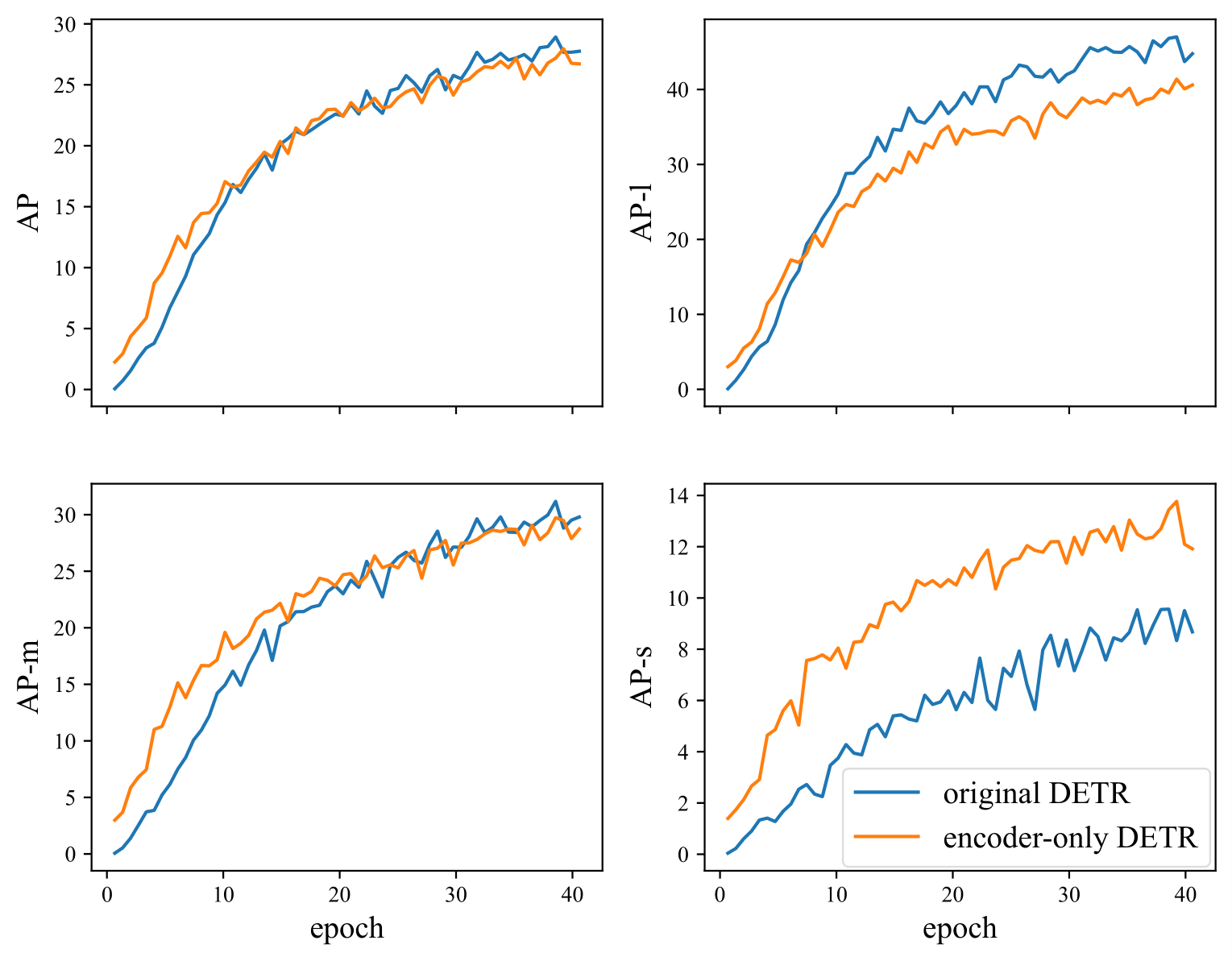}
    \caption{AP, AP-l, AP-m, and AP-s results on COCO validation set: original DETR v.s. encoder-only DETR. We can see that encoder-only DETR significantly accelerate the training of small object detection.}
    \label{fig:encoder}
\end{figure}

Figure~\ref{fig:encoder} presents the the Average Precision (AP) curves of the original DETR and the encoder-only DETR, including the 
overall AP curve (denoted as AP) and the curves for large (AP-l), medium (AP-m), and small (AP-s) objects\footnote{We follow the definitions of small, medium, and large objects in \cite{lin2014microsoft}.}, respectively.
The over-all curves (left upper corner) show that the encoder-only DETR 
performs as well as the original DETR. This means that we can remove the cross-attention part from DETR
without much performance degeneration, which is a positive result. %
From the remaining curves we can see that the encoder-only DETR outperforms the original DETR significantly on small objects and partly on medium object, but under-performs on large objects on the other hand.
A potential interpretation, we think, is that a large object may include too many potentially matchable feature points, which are difficult for the sliding point scheme in the encoder-only DETR to handle.
Another possible reason is that a single feature map processed by encoder is not robust for %
predicting objects of different \mbox{scales \cite{lin2017feature}}.

\begin{figure*}
    \centering
    \includegraphics[width=0.7\textwidth]{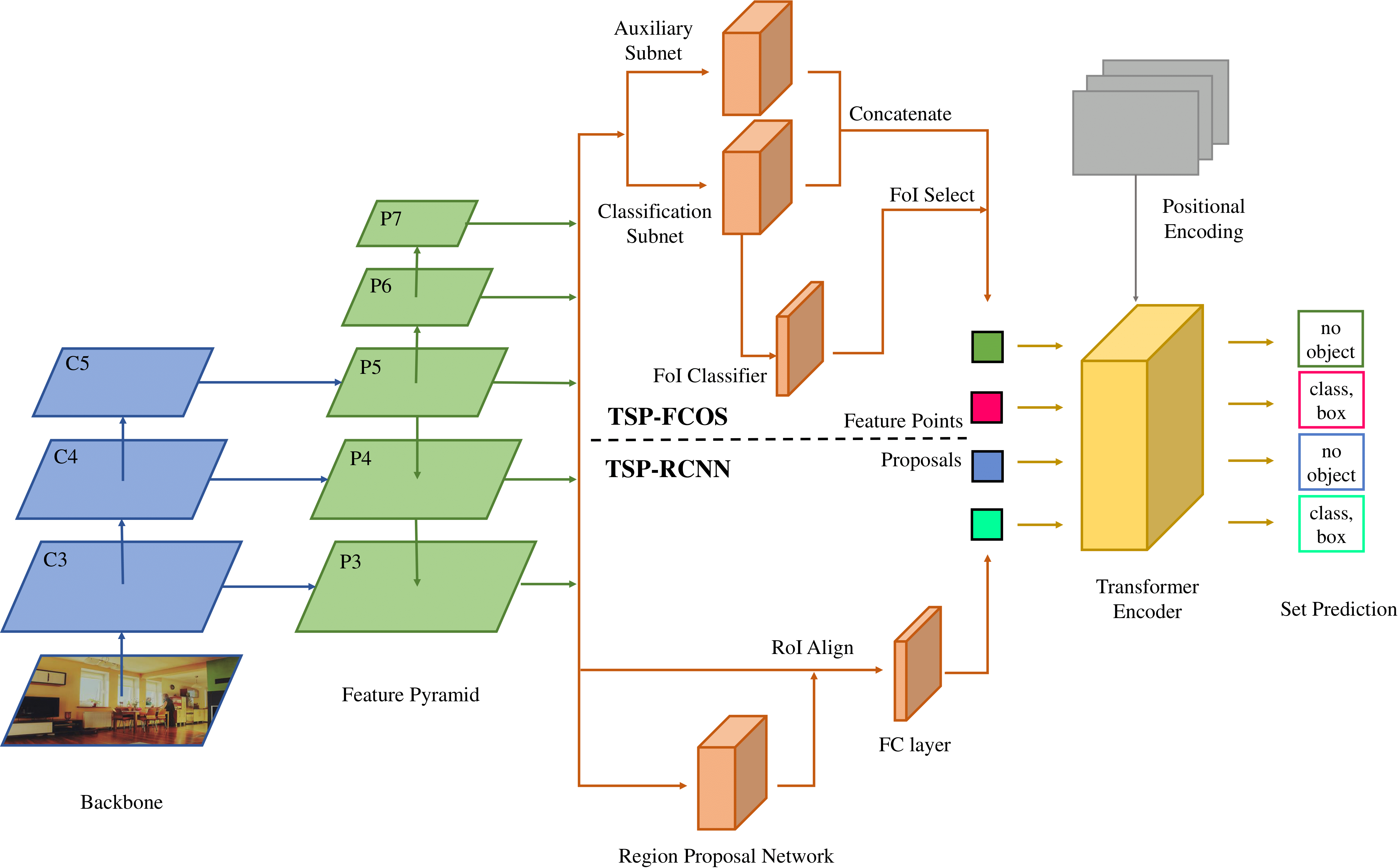}
    \caption{The network architectures of TSP-FCOS and TSP-RCNN, where $C_3$ to $C_5$ denote the feature maps of the backbone network and $P_3$ to $P_7$ of the Feature Pyramid Network (FPN). Both TSP-FCOS (upper) and TSP-RCNN (lower) are equipped with a Transformer encoder and trained with set prediction loss. The difference between them is that the FoI classifier in TSP-FCOS only predicts objectness of each feature (\ie, Features of Interest), but the Region Proposal Network (RPN) in TSP-RCNN predicts both bounding boxes and their objectness as Regions of Interest (RoI), namely, proposals.}
    \label{fig:main}
\end{figure*}

\section{The Proposed Methods} \label{sec:our_method}

Based on our analysis in the previous section, for speeding up the convergence of DETR we need to address both the instability issue in the bipartite matching part of DETR and the cross-attention issue in Transformer modules. Specifically, in order to leverage the speed-up potential of encoder-only DETR we need to overcome its weakness in handling the various scales of objects.
Recently, FCOS \cite{tian2019fcos} (Fully Convolutional One-Stage object detector) shows that multi-level prediction with Feature Pyramid Network (FPN) \cite{lin2017feature} is a good solution to this problem. Inspired by this work %
we propose our first model, namely Transformer-based Set Prediction with FCOS (TSP-FCOS). Then based on TSP-FCOS, we further apply two-stage refinement, which leads to our second model, namely, %
Transformer-based Set Prediction with RCNN (TSP-RCNN).

\subsection{TSP-FCOS}

TSP-FCOS combines the strengths of both FCOS and encoder-only DETR, with a novel component namely Feature of Interest (FoI) selection which enables the Transformer encoder to handle multi-level features,
and a new bipartite matching scheme %
for faster set prediction training. Figure~\ref{fig:main} (upper part) illustrates the network architecture of TSP-FCOS, with the following components:

    \vspace{-10pt}

    \paragraph{Backbone and FPN} We follow FCOS \cite{tian2019fcos} on the design of the backbone and the Feature Pyramid Network (FPN) \cite{lin2017feature}.
    At the beginning of the pipeline, a backbone CNN is used to extract features from the input images. Based on the feature maps from the backbone, we build %
    the FPN component which produces %
    multi-level features that can help encoder-only DETR detect objects of various scales.

    \vspace{-10pt}

    \paragraph{Feature extraction subnets} For a fair comparison with other one-stage detectors (\eg, FCOS and RetinaNet), we follow their design and use two feature extraction heads shared across different feature pyramid levels. We call one of them classification subnet (head), which is used for FoI classification. The other is called auxiliary subnet (head). Their outputs are concatenated and then selected by FoI classifier.

    \vspace{-10pt}

    \paragraph{Feature of Interest (FoI) classifier} In the self-attention module of Transformer, the computation complexity is quadratic to the sequence length, which prohibits directly using all the features on the feature pyramids. To improve the efficiency of self-attention, we design a binary classifier to select a limited portion of features and refer them as Features of Interest (FoI). The binary FoI classifier is trained with FCOS's ground-truth assignment rule\footnote{Please refer to the FCOS paper \cite{tian2019fcos} for more details.}. After FoI classification, top scored features are picked as FoIs and fed into the Transformer encoder.

    \vspace{-10pt}

    \paragraph{Transformer encoder} 
    After the FoI selection step, the input to Transformer encoder is a set of FoIs and their corresponding positional encoding. Inside each layer of Transformer encoder, self-attention is performed to aggregate the information of different FoIs. The outputs of the encoder pass through a shared feed forward network, which predicts the category label (including ``no object'') and bounding box for each FoI.

    \vspace{-10pt}

    \paragraph{Positional encoding} Following DETR, we generalize the positional encoding of Transformer \cite{vaswani2017attention} to the 2D image scenario. Specifically, for a feature point with normalized position $(x, y)  \in [0, 1]^2$, its positional encoding is defined as $[PE(x) : PE(y)]$, where $[:]$ denotes concatenation and function $PE$ is defined by:
    \begin{equation}
        \begin{split}
        PE(x)_{2i} &= \sin(x / 10000^{2i/\text{d}_{\text{model}}})\\
        PE(x)_{2i + 1} &= \cos(x / 10000^{2i/\text{d}_{\text{model}}})
        \end{split}
    \end{equation}
    where $\text{d}_{\text{model}}$ is the dimension of the FoIs.

    \paragraph{Faster set prediction training} As mentioned in Section~\ref{sec:set prediction}, the object detection task can be viewed as a set prediction problem. Given the set of detection results and ground truth objects, the set prediction loss links them together and provides an objective for the model to optimize. But as we show in Section~\ref{sec:biparitite}, the Hungarian bipartite-matching loss can lead to slow convergence in the early stage of training. Therefore, we design a new bipartite matching scheme for faster set prediction training of TSP-FCOS. Specifically, a feature point can be assigned to a ground-truth object only when the point is in the bounding box of the object and in the proper feature pyramid level. This is inspired by the ground-truth assignment rule of FCOS \cite{tian2019fcos}. Next, a restricted cost-based matching process (Equation~\ref{eq:match}) is performed to determine the optimal matching between the detection results and the ground truth objects in the Hungarian loss (Equation~\ref{eq:loss}).

\subsection{TSP-RCNN}

Based on the design of TSP-FCOS and Faster RCNN, we can combine the best of them and perform a two-stage bounding box refinement as set prediction, which requires more computational resources but can detect objects more accurately. This idea leads to TSP-RCNN (Transformer-based Set Prediction with RCNN). Figure~\ref{fig:main} (lower part) illustrates the network architecture of our proposed TSP-RCNN. The main differences between TSP-FCOS and TSP-RCNN are as follows:

    \vspace{-5pt}

    \paragraph{Region proposal network} In TSP-RCNN, instead of using two feature extraction heads and FoI classifier to obtain the input of Transformer encoder, we follow the design of Faster RCNN \cite{ren2015faster} and use a Region Proposal Network (RPN) to get a set of Regions of Interest (RoIs) to be further refined. Different from FoIs in TSP-FCOS, each RoI in TSP-RCNN contains not only an objectness score, but also a predicted bounding box. We apply RoIAlign \cite{he2017mask} to extract the information of RoIs from multi-level feature maps. The extracted features are then flattened and fed into a fully connected network as the input of Transformer encoder.

    \vspace{-5pt}

    \paragraph{Positional encoding} The positional information of a RoI (proposal) is defined by four quantities $(cx, cy, w, h)$, where $(cx, cy) \in [0, 1]^2$ denotes the normalized center coordinates and $(w, h) \in [0, 1]^2$ denotes the normalized height and width. We use $[PE(cx) : PE(cy) : PE(w) : PE(h)]$ as the positional encoding of the proposal, where $PE$ and $[:]$ is defined in the same way as TSP-FCOS.

    \vspace{-5pt}

    \paragraph{Faster set prediction training} TSP-RCNN is also trained with a set prediction loss. Different from TSP-FCOS, we borrow the ground-truth assignment rule from Faster RCNN for faster set prediction training of TSP-RCNN. Specifically, a proposal can be assigned to a ground-truth object if and only if the intersection-over-union (IoU) score between their bounding boxes is greater than 0.5.

\begin{table*}[!ht]
\small
\begin{center}
\begin{tabular}{l | c|c | c c c c c c | c c}
\toprule
Model & Backbone & Epochs & AP & $\text{AP}_{\text{50}}$ & $\text{AP}_{\text{75}}$ & $\text{AP}_{\text{S}}$ & $\text{AP}_{\text{M}}$ & $\text{AP}_{\text{L}}$ & FLOPs & FPS\\
\midrule
FCOS\dag &  ResNet-50 & 36 & 41.0 & 59.8 & 44.1 & 26.2 & 44.6 & 52.2 & 177G & 17\\
Faster RCNN-FPN & ResNet-50 & 36 & 40.2 & 61.0 & 43.8 & 24.2 & 43.5 & 52.0 & 180G & 19\\
Faster RCNN-FPN+ & ResNet-50 & 108 & 42.0 & 62.1 & 45.5 & 26.6 & 45.4 & 53.4 & 180G & 19\\
DETR+ & ResNet-50 & 500 & 42.0 & 62.4 & 44.2 & 20.5 & 45.8 & 61.1 & 86G & 21\\
DETR-DC5+ & ResNet-50 & 500 & 43.3 & 63.1 & 45.9 & 22.5 & 47.3 & 61.1 & 187G & 7\\
Deformable DETR$^*$ & ResNet-50 & 50 & 43.8 & 62.6 & 47.7 & 26.4 & 47.1 & 58.0 & 173G & -\\
UP-DETR & ResNet-50 & 300 & 42.8 & 63.0 & 45.3& 20.8 & 47.1 & \textbf{61.7} & 86G & 21\\
\midrule
TSP-FCOS & ResNet-50 & 36 & 43.1 & 62.3 & 47.0 & 26.6 & 46.8 & 55.9 & 189G & 15\\
TSP-RCNN & ResNet-50 & 36 & 43.8 & 63.3 & 48.3 & 28.6 & 46.9 & 55.7 & 188G & 11\\
TSP-RCNN+ & ResNet-50 & 96 & \textbf{45.0} & \textbf{64.5} & \textbf{49.6} & \textbf{29.7} & \textbf{47.7} & 58.0 & 188G & 11\\
\midrule
FCOS\dag &  ResNet-101 & 36 & 42.5 & 61.3 & 45.9 & 26.0 & 46.5 & 53.6 & 243G & 13\\
Faster RCNN-FPN & ResNet-101 & 36 & 42.0 & 62.5 & 45.9 & 25.2 & 45.6 & 54.6 & 246G & 15\\
Faster RCNN-FPN+ & ResNet-101 & 108 &  44.0 & 63.9 & 47.8 & 27.2 & 48.1 & 56.0 & 246G & 15\\
DETR+ & ResNet-101 & 500 & 43.5 & 63.8 & 46.4 & 21.9 & 48.0 & 61.8 & 152G & 15\\
DETR-DC5+ & ResNet-101 & 500 & 44.9 & 64.7 & 47.7 & 23.7 & 49.5 & \textbf{62.3} &  253G & 6\\
\midrule
TSP-FCOS & ResNet-101 & 36 & 44.4 & 63.8 & 48.2 & 27.7 & 48.6 & 57.3 & 255G & 12\\
TSP-RCNN & ResNet-101 & 36 & 44.8 & 63.8 & 49.2 & 29.0 & 47.9 & 57.1 & 254G & 9\\
TSP-RCNN+ & ResNet-101 & 96 & \textbf{46.5} & \textbf{66.0} & \textbf{51.2} & \textbf{29.9} & \textbf{49.7} & 59.2 & 254G & 9\\
\bottomrule
\end{tabular}
\end{center}
\caption{Evaluation results on COCO 2017 validation set. \dag\ represents our reproduction results. + represents that the models are trained with random crop augmentation and a longer training schedule. We use Detectron2 package to measure FLOPs and FPS. A single Nvidia GeForce RTX 2080 Ti GPU is used for measuring inference latency. $*$ represents the version without iterative refinement. A fair comparison of TSP-RCNN and Deformabel DETR both with iterative refinement can be found in the appendix.}
\label{tab:main}
\end{table*}

\section{Experiments}

\subsection{Dataset and Evaluation Metrics}
We evaluate our methods on the COCO \cite{lin2014microsoft} object detection dataset, which includes 80 object classes.
Following the common practice \cite{lin2017focal,tian2019fcos}, we use all $115k$ images in \texttt{trainval35k} split for training and all $5k$ images in \texttt{minival} split for validation. The test result is obtained by submitting the results of \texttt{test-dev} split to the evaluation server.
For comparison with other methods, we mainly focus on the Average Precision (AP), which is the primary challenge metric used in COCO, and FLOPs, which measures the computation overhead.

\subsection{Implementation Details}
We briefly describe the default settings of our implementation. More detailed settings can be found in appendix.

    \vspace{-5pt}

    \paragraph{TSP-FCOS} Following FCOS \cite{tian2019fcos}, both classification subnet and auxiliary subnet use four $3\times3$ convolutional layers with 256 channels and group normalization \cite{wu2018group}. In FoI selection, we select top 700 scored feature positions from FoI classifier as the input of Transformer encoder.

    \vspace{-5pt}

    \paragraph{TSP-RCNN} Different from the original Faster RCNN, we apply 2 unshared convolutional subnets to $P_3$-$P_7$ as classification and regression heads of RPN and use a RetinaNet \cite{lin2017focal} style anchor generation scheme. We find this improves the performance of RPN with less computation overhead. In RoI selection, we select top 700 scored features from RPN. RoI Align operation \cite{he2017mask} and a fully connected layer are applied to extract the proposal features from RoIs.

    \vspace{-5pt}

    \paragraph{Transformer encoder} As both TSP-FCOS and TSP-RNN only have a Transformer encoder while DETR has both Transformer encoder and decoder, to be comparable in terms of FLOPs with DETR-DC5, we use a 6-layer Transformer encoder of width 512 with 8 attention heads. The hidden size of feed-forward network (FFN) in Transformer is set to 2048. During training, we randomly drop 70\% inputs of Transformer encoder to improve the robustness of set prediction.

    \vspace{-5pt}

    \paragraph{Training} We follow the default setting of Detectron2 \cite{wu2019detectron2}, where a 36-epoch ($3\times$) schedule with multi-scale train-time augmentation is used.

\begin{table}[!t]
\small
\begin{center}
\begin{tabular}{l | c c c c}
\toprule
Model  & AP & $\text{AP}_{\text{S}}$ & $\text{AP}_{\text{M}}$ & $\text{AP}_{\text{L}}$\\
\midrule
TSP-RCNN-R50 & \textbf{43.8} & \textbf{28.6} & \textbf{46.9} & 55.7\\
\quad w/o set prediction loss & 42.7 & 27.6 & 45.5 & \textbf{56.2}\\
\quad w/o positional encoding & 43.4 & 28.4 & 46.3 & 55.0\\
TSP-RCNN-R101 & \textbf{44.8}  & \textbf{29.0} & \textbf{47.9} & \textbf{57.1}\\
\quad w/o set prediction loss & 44.0 & 27.6 & 47.2 & \textbf{57.1}\\
\quad w/o positional encoding & 44.4 & 28.2 & 47.7 & 56.7\\
\bottomrule
\end{tabular}
\end{center}
\caption{Evaluation results on COCO 2017 validation set for ablation study of set prediction loss and positional encoding.}
\label{tab:ablation}
\end{table}

\begin{table}[!t]
\small
\begin{center}
\begin{tabular}{l | c c c c}
\toprule
Model & AP & $\text{AP}_{\text{S}}$ & $\text{AP}_{\text{M}}$ & $\text{AP}_{\text{L}}$\\
\midrule
FCOS & 45.3 & 28.1 & 49.0 & 59.3 \\
TSP-FCOS & \textbf{46.1} & \textbf{28.5} & \textbf{49.7} & \textbf{60.2}\\
\midrule
Faster-RCNN & 44.1 & 26.4 & 47.6 & 58.1\\
TSP-RCNN & \textbf{45.8} & \textbf{29.4} & \textbf{49.2} & \textbf{58.4}\\
\bottomrule
\end{tabular}
\end{center}
\caption{Evaluation results on COCO 2017 validation set with ResNet-101-DCN backbone.}
\label{tab:dcn}
\end{table}

\begin{table*}[!t]
\small
\begin{center}
\begin{tabular}{l | c | c c c c c c }
\toprule
Model & Backbone & AP & $\text{AP}_{\text{50}}$ & $\text{AP}_{\text{75}}$ & $\text{AP}_{\text{S}}$ & $\text{AP}_{\text{M}}$ & $\text{AP}_{\text{L}}$\\
\midrule
RetinaNet \cite{lin2017focal} & ResNet-101 & 39.1 & 59.1 & 42.3 & 21.8 & 42.7 & 50.2\\
FSAF \cite{zhu2019feature} & ResNet-101 & 40.9 & 61.5 & 44.0 & 24.0 & 44.2 & 51.3\\
FCOS \cite{tian2019fcos} & ResNet-101 & 41.5 & 60.7 & 45.0 & 24.4 & 44.8 & 51.6\\
MAL \cite{ke2020multiple} & ResNet-101 & 43.6 & 62.8 & 47.1 & 25.0 & 46.9 & 55.8\\
RepPoints \cite{yang2019reppoints} & ResNet-101-DCN & 45.0 & 66.1 & 49.0 & 26.6 & 48.6 & 57.5 \\
ATSS \cite{zhang2020bridging} & ResNet-101 & 43.6 & 62.1 & 47.4 & 26.1 & 47.0 & 53.6\\
ATSS \cite{zhang2020bridging} & ResNet-101-DCN & 46.3 & 64.7 & 50.4 & 27.7 & 49.8 & 58.4\\
\midrule
Fitness NMS \cite{tychsen2018improving} & ResNet-101 & 41.8 & 60.9 & 44.9 & 21.5 & 45.0 & 57.5\\
Libra RCNN \cite{pang2019libra} & ResNet-101 &  41.1 & 62.1 & 44.7 & 23.4 & 43.7 & 52.5\\
Cascade RCNN \cite{cai2018cascade} & ResNet-101 & 42.8 & 62.1 & 46.3 & 23.7 & 45.5 & 55.2\\
TridentNet \cite{li2019scale} & ResNet-101-DCN & 46.8 & \textbf{67.6} & 51.5 & 28.0 & \textbf{51.2} & \textbf{60.5}\\
TSD \cite{song2020revisiting} & ResNet-101 & 43.2 & 64.0 & 46.9 & 24.0 & 46.3 & 55.8\\
Dynamic RCNN \cite{zhang2020dynamic} & ResNet-101 & 44.7 & 63.6 & 49.1 & 26.0 & 47.4 & 57.2\\
Dynamic RCNN \cite{zhang2020dynamic} & ResNet-101-DCN & 46.9 & 65.9 & 51.3 & 28.1 & 49.6 & 60.0\\
\midrule
TSP-RCNN& ResNet-101 & \underline{46.6} & \underline{66.2} & \underline{51.3} & \underline{28.4} & \underline{49.0} & \underline{58.5}\\
TSP-RCNN& ResNet-101-DCN & \textbf{47.4} & 66.7 & \textbf{51.9} & \textbf{29.0} & 49.7 & 59.1\\
\bottomrule
\end{tabular}
\end{center}
\caption{Comparison with state-of-the-art models on COCO 2017 test set (single-model and single-scale results). \underline{Underlined} and \textbf{bold} numbers represent the best model with ResNet-101 and ResNet-101-DCN backbone, respectively.}
\label{tab:sota}
\end{table*}

\subsection{Main Results}
Table~\ref{tab:main} shows our main results on COCO 2017 validation set. We compare TSP-FCOS and TSP-RCNN with FCOS \cite{tian2019fcos}, Faster RCNN \cite{ren2015faster}, and DETR \cite{carion2020end}. We also compare with concurrent work on improving DETR: Deformable DETR \cite{zhu2020deformable} and UP-DETR \cite{dai2020up}. From the table, we can see that our TSP-FCOS and TSP-RCNN significantly outperform original FCOS and Faster RCNN. Besides, we can find that TSP-RCNN is better than TSP-FCOS in terms of overall performance and small object detection but slightly worse in terms of inference latency.

To compare with state-of-the-art DETR models, we use a similar training strategy in DETR \cite{carion2020end}, where a 96-epoch ($8\times$) training schedule and random crop augmentation is applied. We denote the enhanced version of TSP-RCNN by TSP-RCNN+. We also copy the results of enhanced Faster RCNN (\ie, Faster RCNN+) from \cite{carion2020end}. Comparing these models, we can find that our TSP-RCNN obtains state-of-the-art results with a shorter training schedule. We also find that TSP-RCNN+ still under-performs DETR-DC5+ on large object detection. We think this is because of the inductive bias of the encoder-decoder scheme used by DETR and its longer training schedule.

\subsection{Model Analysis}
For model analysis, we evaluate several models trained in our default setting, \ie, with a 36-epoch ($3\times$) schedule and without random crop augmentation.

\subsubsection{Ablation study}

We conduct an ablation study of set prediction loss and positional encoding, which are two essential components in our model.  Table~\ref{tab:ablation} show the results of ablation study for TSP-RCNN with ResNet-50 and ResNet-101 backbone. From the table, we can see that both set prediction loss and positional encoding are very important to the success of our TSP mechanism, while set prediction loss contributes more than positional encoding to the improvement of TSP-RCNN.

\subsubsection{Compatibility with deformable convolutions}

One may wonder whether Transformer encoder and deformable convolutions \cite{dai2017deformable,zhu2019deformable} are compatible with each other, as both of them can utilize long-range relation between objects. In Table~\ref{tab:dcn}, we compare TSP-FCOS and TSP-RCNN to FCOS and Faster RCNN with deformable ResNet-101 as backbone. From the results, we can see that the TSP mechanism is well complementary with deformable convolutions.

\subsection{Comparison with State-of-the-Arts}

We compare TSP-RCNN with multiple one-stage and two-stage object detection models \cite{ren2015faster, tychsen2018improving, cai2018cascade, song2020revisiting, lin2017focal, zhu2019feature, tian2019fcos, chen2019revisiting, kong2020foveabox, yang2019reppoints, zhang2020bridging} that also use ResNet-101 backbone or its deformable convolution network (DCN) \cite{zhu2019deformable} variant in Table~\ref{tab:sota}.
A $8\times$ schedule and random crop augmentation is used.
The performance metrics are evaluated on COCO 2017 test set using single-model and single-scale detection results. Our model achieves the highest AP scores among all detectors in both backbone settings.

\begin{figure}
    \centering
    \includegraphics[width=0.9\linewidth]{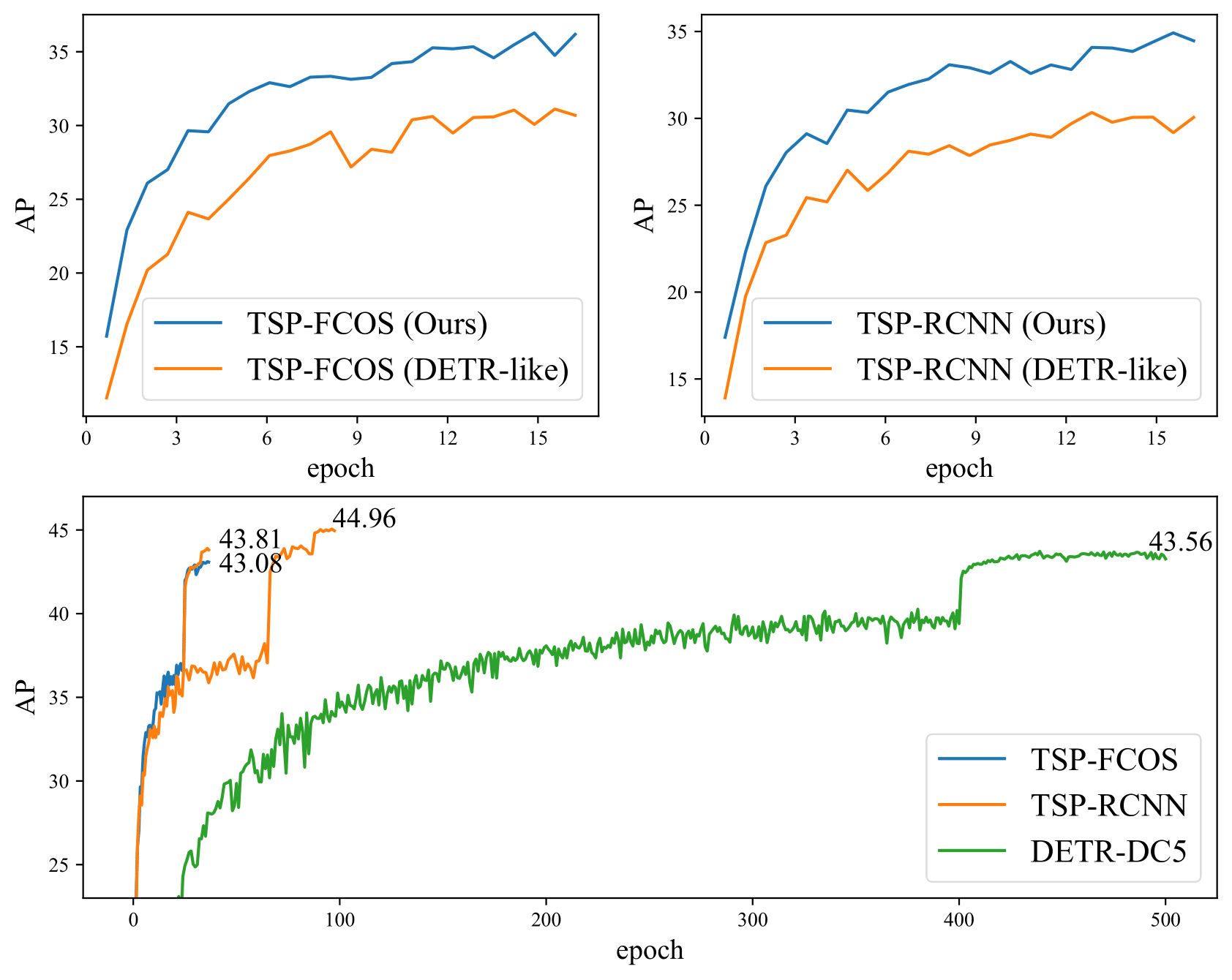}
    \caption{Top two plots compare the convergence speed of our proposed faster set prediction training loss and DETR-like loss for TSP-FCOS and TSP-RCNN. The bottom plot shows the convergence curves of TSP-FCOS, TSP-RCNN, and DETR-DC5.
    }
    \label{fig:convergence}
\end{figure}

\section{Analysis of convergence}

We compare the convergence speed of our faster set prediction training and DETR's original set prediction training in the upper part of Figure~\ref{fig:convergence}. We can see that our proposed faster training technique consistently accelerates the convergence of both TSP-FCOS and TSP-RCNN.

We also plot the convergence curves of TSP-FCOS, TSP-RCNN, and DETR-DC5 in the lower part of Figure~\ref{fig:convergence}, from which we can find that our proposed models not only converge faster, but also achieve better detection performance.

\section{Conclusion}

Aiming to accelerate the training convergence of DETR as well as to improve prediction power in object detection, we present an investigation on the causes of its
slow convergence through extensive experiments, and propose two novel solutions, namely %
TSP-FCOS and TSP-RCNN, which require much less training time and achieve the state-of-the-art detection performance. For future work, we would like to investigate the successful use of sparse attention mechanism \cite{child2019generating,correia2019adaptively,zaheer2020big} for directly modeling the relationship among multi-level features.

\section*{Acknowledgements}
We thank the reviewers for their helpful comments. This work is supported in part by the United States Department of Energy via the Brookhaven National Laboratory under Contract PO 0000384608.

{\small
\bibliographystyle{ieee_fullname}
\bibliography{egbib}
}

\clearpage
\appendix

\section{Preliminaries}

\subsection{Transformer and Detection Transformer}

As this work aims to improve the DEtection TRansformer (DETR) model \cite{carion2020end}, for completeness, we describe its architecture in more details.

\paragraph{Encoder-decoder framework} DETR can be formulated in an encoder-decoder framework \cite{cho2014learning}. The encoder of DETR takes the features processed by the CNN backbone as inputs and generates the context representation, and the non-autoregressive decoder of DETR takes the object queries as inputs and generates the detection results conditional on the context. 

\paragraph{Multi-head attention} Two types of multi-head attentions are used in DETR: multi-head self-attention and multi-head cross-attention. A general attention mechanism can be formulated as the weighted sum of the value vectors $V$ using query vectors $Q$ and key vectors $K$:
\begin{equation}
    \text{Attention}(Q,K,V) = \text{softmax}\left(\frac{QK^T}{\sqrt{d_\text{model}}}\right)\cdot V,
\end{equation}
where $d_\text{model}$ represents the dimension of hidden representations. For self-attention, $Q$, $K$, and $V$ are hidden representations of the previous layer. For cross-attention, $Q$ refers to hidden representations of the previous layer, whereas $K$ and $V$ are context vectors from the encoder. Multi-head variant of the attention mechanism allows the model to jointly attend to information from different representation subspaces, and is defined as:
\begin{equation*}
    \begin{split}
        &\text{Multi-head}(Q, K, V) = \text{Concat}(\text{head}_1, \dots, \text{head}_H) W^O,\\
        &\text{head}_h = \text{Attention}((P_Q + Q)W^Q_h,(P_K+K)W^K_h,VW^V_h),
    \end{split}
\end{equation*}
where $W^Q_h, W^K_h \in \mathbb{R}^{d_\text{model}\times d_k}$, $W^V_h \in \mathbb{R}^{d_\text{model}\times d_v}$, and $W^O_h \in \mathbb{R}^{Hd_v \times d_\text{model}}$ are projection matrices, $H$ is the number of attention heads, $d_k$ and $d_v$ are the hidden sizes of queries/keys and values per head, and $P_Q$ and $P_K$ are positional encoding.

\paragraph{Feed-forward network} The position-wise Feed-Forward Network (FFN) is applied after multi-head attentions in both encoder and decoder. It consists of a two-layer linear transformation with ReLU activation:
\begin{equation}
    \text{FFN}(x) = \max(0, xW_1 + b_1) W_2 + b_2,
\end{equation}
where $W_1 \in \mathbb{R}^{d_\text{model} \times d_\text{FFN}}$, $W_2 \in \mathbb{R}^{d_\text{FFN} \times d_\text{model}}$, $b_1 \in \mathbb{R}^{d_\text{FFN}}$, $b_2 \in \mathbb{R}^{d_\text{model}}$, and $d_\text{FFN}$ represents the hidden size of FFN.

\begin{figure}
    \centering
    \includegraphics[width=0.8\linewidth]{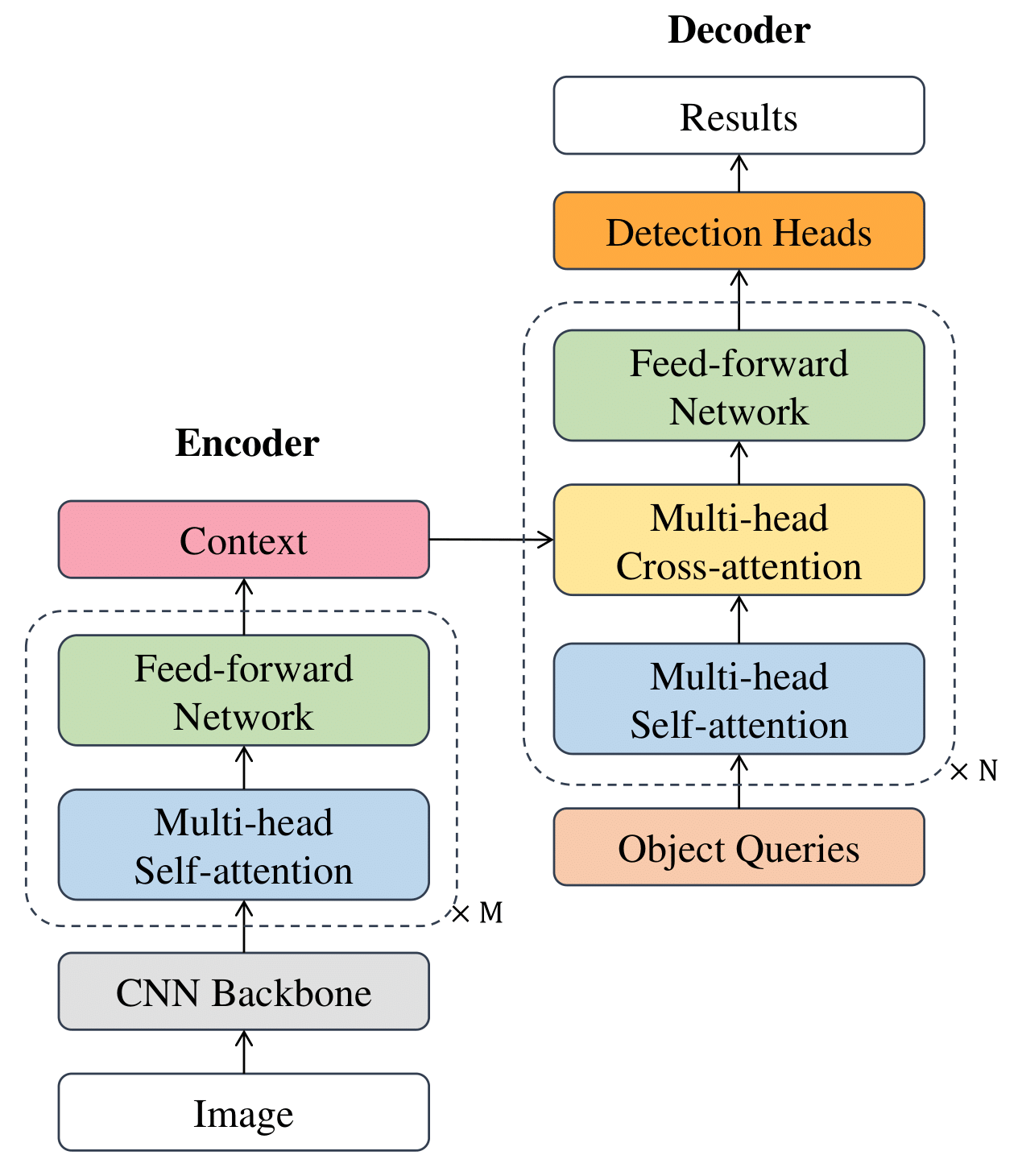}
    \caption{A detailed illustration of the DETR architecture. Residual connection and layer normalization are omitted.}
    \label{fig:detr-orig}
\end{figure}

\paragraph{Stacking} Multi-head attention and feed-forward network are stacked alternately to form the encoder and the decoder, with residual connections \cite{he2016deep} and layer normalization \cite{ba2016layer}. Figure~\ref{fig:detr-orig} shows a detailed illustration of the DETR architecture.

\subsection{Faster R-CNN}

Faster R-CNN \cite{ren2015faster} is a two-stage object detection model, developed based on previous work of R-CNN \cite{girshick2014rich} and Fast R-CNN \cite{girshick2015fast}. With Region Proposal Networks (RPN), Faster R-CNN significantly improves the accuracy and efficiency of two-stage object detection.

\paragraph{Region Proposal Networks} The first module of Faster R-CNN is a deep fully convolutional network, named the Region Proposal Network (RPN) that proposes Regions of Interest (RoIs). RPN takes the feature maps of a image as input, and outputs a set of rectangular object proposals with their objectness scores. RPN contains a shared $3\times 3$ convolutional layer and two sibling $1\times 1$ convolutional layers for regression and classification respectively. At each sliding-window location, RPN produces $k$ proposals. The proposals are parameterized relative to $k$ reference boxes called anchors. In Fast R-CNN, $3$ scales and $3$ aspect ratios of anchors are used, so there are $k=9$ anchors for each sliding window. For each anchor, the regression head outputs $4$ coordinate parameters $\{t_x,t_y,t_w,t_h\}$ that encode the location and size of bounding boxes, and the classification head outputs $2$ scores $\{p_\text{pos}, p_\text{neg}\}$ that estimate probability of existence of object in the box.

\paragraph{Fast R-CNN} The second part is the Fast R-CNN detector that uses each proposal from RPN to refine the detection. To reduce redundancy, non-maximum suppression (NMS) is applied on the proposals, and only the top ranked proposals can be used by Fast R-CNN. Then, RoI Pooling or RoI Align \cite{he2017mask} is used to extract features from the backbone feature map at the given proposal regions, such that the input to the Fast R-CNN detector has fixed spatial size for each proposal. At this stage, Fast R-CNN outputs bounding box regression parameters and classification scores to refine the region proposals. Again, NMS is required to reduce duplication in the detection results.

\subsection{FCOS}

Fully Convolutional One-Stage Object Detection (FCOS) \cite{tian2019fcos} is a recent anchor-free, per-pixel detection framework that has achieved state-of-the-art one-stage object detection performance.

\paragraph{Per-Pixel Prediction} In contrast to anchor-based object detectors, FCOS formulates the task in a per-pixel prediction fashion, that is, the target bounding boxes are regressed at each location on the feature map, without referencing pre-defined anchors. A location on the feature map is considered as a positive sample if its corresponding position on the input image falls into any ground-truth box. If one location falls into the overlap of multiple ground-truth boxes, the smallest one is selected. Experiments show that with multi-level prediction and FPN \cite{lin2017feature}, this ambiguity does not affect the overall performance.

\paragraph{Network Outputs} In FCOS, there are two branches after the feature maps from the backbone. The first branch has $4$ convolutional layers and two sibling layers that outputs $C$ classification scores and a ``center-ness'' score. The center-ness depicts the normalized distance from the location to the center of the object that the location is responsible for. The center-ness ranges in $[0, 1]$ and is trained with binary cross entropy loss. During test, the center-ness is multiplied to the classification score, thus the possibly low-quality bounding boxes that are far away from the center of objects will have less weight in NMS. The second branch has $4$ convolutional layers and a bounding box regression layer that outputs the distance from the location to the four sides of the box. The prediction head is shared across multiple feature levels.

\begin{table*}[!ht]
\small
\begin{center}
\begin{tabular}{l | c | c c c c c c }
\toprule
Model & Backbone & AP & $\text{AP}_{\text{50}}$ & $\text{AP}_{\text{75}}$ & $\text{AP}_{\text{S}}$ & $\text{AP}_{\text{M}}$ & $\text{AP}_{\text{L}}$\\
\midrule
Faster RCNN \cite{ren2015faster} & ResNet-101 & 36.2 & 59.1 & 39.0 & 18.2 & 39.0 & 48.2\\
Fitness NMS \cite{tychsen2018improving} & ResNet-101 & 41.8 & 60.9 & 44.9 & 21.5 & 45.0 & 57.5\\
Libra RCNN \cite{pang2019libra} & ResNet-101 &  41.1 & 62.1 & 44.7 & 23.4 & 43.7 & 52.5\\
Cascade RCNN \cite{cai2018cascade} & ResNet-101 & 42.8 & 62.1 & 46.3 & 23.7 & 45.5 & 55.2\\
TridentNet \cite{li2019scale} & ResNet-101-DCN & 46.8 & 67.6 & 51.5 & 28.0 & 51.2 & 60.5\\
TSD \cite{song2020revisiting} & ResNet-101 & 43.2 & 64.0 & 46.9 & 24.0 & 46.3 & 55.8\\
Dynamic RCNN \cite{zhang2020dynamic} & ResNet-101 & 44.7 & 63.6 & 49.1 & 26.0 & 47.4 & 57.2\\
Dynamic RCNN \cite{zhang2020dynamic} & ResNet-101-DCN & 46.9 & 65.9 & 51.3 & 28.1 & 49.6 & 60.0\\
\midrule
RetinaNet \cite{lin2017focal} & ResNet-101 & 39.1 & 59.1 & 42.3 & 21.8 & 42.7 & 50.2\\
FSAF \cite{zhu2019feature} & ResNet-101 & 40.9 & 61.5 & 44.0 & 24.0 & 44.2 & 51.3\\
FCOS \cite{tian2019fcos} & ResNet-101 & 41.5 & 60.7 & 45.0 & 24.4 & 44.8 & 51.6\\
MAL \cite{ke2020multiple} & ResNet-101 & 43.6 & 62.8 & 47.1 & 25.0 & 46.9 & 55.8\\
RepPoints \cite{yang2019reppoints} & ResNet-101-DCN & 45.0 & 66.1 & 49.0 & 26.6 & 48.6 & 57.5 \\
ATSS \cite{zhang2020bridging} & ResNet-101 & 43.6 & 62.1 & 47.4 & 26.1 & 47.0 & 53.6\\
ATSS \cite{zhang2020bridging} & ResNet-101-DCN & 46.3 & 64.7 & 50.4 & \textbf{27.7} & \textbf{49.8} & 58.4\\
\midrule
TSP-FCOS & ResNet-101 & \underline{46.1} & \underline{65.8} & \underline{50.3} & \underline{27.3} & \underline{49.0} & \underline{58.2} \\
TSP-FCOS & ResNet-101-DCN & \textbf{46.8} & \textbf{66.4} & \textbf{51.0} & 27.6 & 49.5 & \textbf{59.0} \\
\bottomrule
\end{tabular}
\end{center}
\caption{Compare TSP-FCOS with state-of-the-art models on COCO 2017 test set (single-model and single-scale results). \underline{Underlined} and \textbf{bold} numbers represent the best one-stage model with ResNet-101 and ResNet-101-DCN backbone, respectively.}
\label{tab:sota2}
\end{table*}

\setlength{\tabcolsep}{3pt}
\begin{table}[!t]
\small
\begin{center}
\begin{tabular}{l | c c c c | c c}
\toprule
Model & AP & $\text{AP}_{\text{S}}$ & $\text{AP}_{\text{M}}$ & $\text{AP}_{\text{L}}$ & FLOPs & \#Params\\
\midrule
FCOS & 41.0 & 26.2 & 44.6 & 52.2 & 177G & 36.4M\\
FCOS-larger & 41.5 & 26.0 & 45.2 & 52.3 & 199G & 37.6M\\
TSP-FCOS & \textbf{43.1} & \textbf{26.6} & \textbf{46.8} & \textbf{55.9} & 189G & 51.5M\\
\midrule
Faster RCNN & 40.2 & 24.2 & 43.5 & 52.0 & 180G & 41.7M\\
Faster RCNN-larger & 40.9& 24.4 & 44.1 & 54.1 & 200G & 65.3M\\
TSP-RCNN & \textbf{43.8} & \textbf{28.6} & \textbf{46.9} & \textbf{55.7} & 188G & 63.6M\\
\bottomrule
\end{tabular}
\end{center}
\caption{Evaluation results on COCO 2017 validation set of models under various FLOPs. ResNet-50 is used as backbone.}
\label{tab:more}
\end{table}
\setlength{\tabcolsep}{6pt}

\section{Detailed Experimental Settings}

We provide more details about the default settings of our implementation.

    \paragraph{Backbone} We use ResNet-50 and ResNet-101 \cite{he2016deep} as the backbone, and a Feature Pyramid Network \cite{lin2017feature} is built on the $\{C_3, C_4, C_5\}$ feature maps from ResNet to produce feature pyramid $\{P_3,P_4,P_5,P_6,P_7\}$. If specified with DCN, we use Deformable ConvNets v2 \cite{zhu2019deformable} in the last three stages of ResNet. The feature maps have 256 channels.

    \paragraph{Data augmentation} We follow the default setting of Detectron2 \cite{wu2019detectron2} for data augmentation. Specifically, we use scale augmentation to resize the input images such that the shortest side is in $\{640, 672, 704, 736, 768, 800\}$, and the longest is no larger than $1333$. Besides scale augmentation, we also randomly flip training images horizontally.

    \paragraph{Loss} We use our proposed faster set prediction training loss for classification, and a combination of L1 and Generalized IoU \cite{rezatofighi2019generalized} losses for regression. Focal loss \cite{lin2017focal} is used for weighting positive and negative examples in classification for both TSP-FCOS and TSP-RCNN. Unlike DETR \cite{carion2020end}, we do not apply auxiliary losses after each encoder layer. We find this end-to-end scheme improves the model performance.

    \paragraph{Optimization} We use AdamW \cite{loshchilov2017decoupled} to optimize the Transformer component, and SGD with momentum $0.9$ to optimizer the other parts in our detector. For the $36$-epoch ($3\times$) schedule, we train the detector for $2.7\times 10^5$ iterations with batch size $16$. The learning rate is set to $10^{-4}$ for AdamW, and $10^{-2}$ for SGD in the beginning, and both multiplied by $0.1$ at $1.8\times 10^5$ and $2.4\times 10^5$ iterations. We also use linear learning rate warm-up in the first $1000$ iterations. The weight decay is set to $10^{-4}$. We apply gradient clipping for the Transformer part, with a maximal $L_2$ gradient norm of $0.1$. 
    
    \paragraph{Longer training schedule} We also use a 96-epoch ($8\times$) schedule in the paper. The 96-epoch ($8\times$) schedule will resume from $36$-epoch ($3\times$) schedule's model checkpoint in the $24^{th}$ epoch (i.e., $1.8 \times 10^5$ iterations), and continue training for $72$ epoch (i.e., $5.4 \times 10^5$ iterations).
    The learning rate is multiplied by $0.1$ at $4.8\times 10^5$ and $6.4\times 10^5$ iterations.
    In the $8\times$ schedule, we will further apply random crop augmentation. We follow the augmentation strategy in DETR \cite{carion2020end}, where a train image is cropped with probability 0.5 to a random rectangular patch which is then resized again to 800-1333. 

\setlength{\tabcolsep}{3pt}
\begin{table}[!t]
\small
\begin{center}
\begin{tabular}{l | c c c c c c}
\toprule
Model & AP & $\text{AP}_{\text{50}}$ & $\text{AP}_{\text{75}}$ & $\text{AP}_{\text{S}}$ & $\text{AP}_{\text{M}}$ & $\text{AP}_{\text{L}}$\\
\midrule
Deformable DETR & 43.8 & 62.6 & 47.7 & 26.4 & 47.1 & 58.0\\
+ iterative refinement & 45.4 & 64.7 & 49.0 & 26.8 & 48.3 & 61.7 \\
++ two-stage$^*$ & \textbf{46.2} & \textbf{65.2} & \textbf{50.0} & 28.8 & \textbf{49.2} & \textbf{61.7} \\
\midrule
TSP-RCNN & 44.4 & 63.7 & 49.0 & 29.0 & 47.0 & 56.7\\
+ iterative refinement & 45.4 & 63.1 & 49.6 & \textbf{29.5} & 48.5 & 58.7\\
\bottomrule
\end{tabular}
\end{center}
\caption{Evaluation results on COCO 2017 validation set of TSP-RCNN and Deformable DETR with iterative refinement. All models are trained with 50 epochs and a batch size of 32. $^*$ Please refer to the original Deformable DETR paper for the definition of two-stage Deformable DETR.}
\label{tab:cascade}
\end{table}
\setlength{\tabcolsep}{6pt}

\section{More Details of Encoder-only DETR}

Our encoder-only DETR is also trained with the Hungarian loss for set prediction, but the bounding box regression process is a bit different.
In original DETR, bounding box regression is reference-free, where DETR directly predicts the normalized center coordinates $(cx, cy) \in [0, 1]^2$, height and width $(w, h) \in [0, 1]^2$ of the box w.r.t. the input image. In encoder-only DETR, as each prediction is based on a feature point of Transformer encoder output, we will use the feature point coordinates $(x_r, y_r)$ as the reference point of regression:
\begin{equation*}
    cx = \sigma(b_1 + \sigma^{-1}(x_r)), cy = \sigma(b_2 + \sigma^{-1}(y_r))
\end{equation*}
where $\{b_1,b_2\}$ are from the output of regression prediction.

\section{Comparison between TSP-RCNN and Deformable DETR with Iterative Refinement}

Inspired by Deformable DETR \cite{zhu2020deformable}, we conduct experiments of TSP-RCNN which also iteratively refines the prediction boxes in a cascade style \cite{cai2018cascade}. Here we implement a simple two-cascade scheme, whether the dimension of fully connected detection head and Transformer feed-forward network are reduced from 12544-1024-1024 and 512-2048-512 to 12544-512 and 512-1024-512, respectively, to maintain a similar number of parameters and FLOPs as the original model. To make a fair comparison, we also follow the experimental setting of Deformable DETR where a 50-epoch training schedule with batch size 32 is used.

Table \ref{tab:cascade} shows the results of TSP-RCNN and Deformable DETR with iterative refinement. From the results, we can see that without iterative refinement, TSP-RCNN outperforms Deformable DETR with the same training setting. The iterative refinement process can improve the performance of TSP-RCNN by 1 AP point. We can also find that both with iterative refinement, TSP-RCNN slightly underperforms Deformable DETR. We believe this is because Deformable DETR utilizes $D=6$ decoder refinement iterations, while we only conduct experiments with two refinement iterations. How to efficiently incorporate multiple refinement iterations into the TSP-RCNN model is left as future work.

\section{Comparison under similar FLOPs}

Compared to original FCOS and Faster RCNN, our TSP-FCOS and TSP-RCNN use an additional Transformer encoder module. Therefore, it is natural to ask whether the improvements come from more computation
and parameters. Table~\ref{tab:more} answers this question by applying stronger baseline models to the baseline models. For Faster RCNN, we first apply two unshared convolutional layers to $P_3$-$P_7$ as a stronger RPN, and then change the original 12544-1024-1024 fully-connected (fc) detection head to 12544-2048-2048-2048. This results in a Faster RCNN model with roughly 200 GFLOPs and 65.3M parameters. For FCOS, we evaluate a FCOS model with roughly 199 GFLOPs, where we add one more convolutional layer in both classification and regression heads. From Table~\ref{tab:more}, we can see that while adding more computation and parameters to baselines can slightly improve their performance, such improvements are not as significant as our TSP mechanism.

\section{Compare TSP-FCOS with State-of-the-Arts}

For completeness, we also compare our proposed TSP-FCOS model with other state-of-the-art detection models \cite{ren2015faster, tychsen2018improving, cai2018cascade, song2020revisiting, lin2017focal, zhu2019feature, tian2019fcos, chen2019revisiting, kong2020foveabox, yang2019reppoints, zhang2020bridging} that also use ResNet-101 backbone or its deformable convolution network (DCN) \cite{zhu2019deformable} variant in Table~\ref{tab:sota2}.
A $8\times$ schedule and random crop augmentation is used.
The performance metrics are evaluated on COCO 2017 test set using single-model and single-scale detection results. We can see that TSP-FCOS achieves state-of-the-art performance among one-stage detectors in terms of the AP score. But comparing Table 4 in the main paper and Table~\ref{tab:sota2}, we can also find that TSP-FCOS slightly under-performs our proposed TSP-RCNN model.

\section{Ablation Study for the Number of Feature Positions \& Proposals}

For TSP-FCOS, we select top 700 scored feature positions from FoI classifier as the input of Transformer encoder during FoI selection, while for for TSP-RCNN, we select top 700 scored proposals from RPN during RoI selection. However, the number of feature Positions and proposals used in our experiments are not necessarily optimal. We present an ablation study with respect to this point in Table \ref{tab:ab_features}. Our results show that our models still preserve a high prediction accuracy when only using half of feature positions.

\begin{table}[t]
    \centering
    \caption{Ablation result w.r.t. the number of proposals for R-50 TSP-RCNN and the number of feature positions for R-50 TSP-FCOS on the validation set.}
    \vspace{0.5em}
    \begin{tabular}{c c c c c}
    \toprule
         Num. of Proposals & 100 & 300 & 500 & 700 \\
    \midrule
         TSP-RCNN & 40.3 & 43.3 & 43.7 & 43.8\\
         TSP-FCOS & 40.0 & 42.5 & 42.9 & 43.1\\
    \bottomrule
    \end{tabular}
    \label{tab:ab_features}
\end{table}

\section{Qualitative Analysis}

\begin{figure*}[t]
\begin{center}
\subfigure{\includegraphics[height=0.3\linewidth]{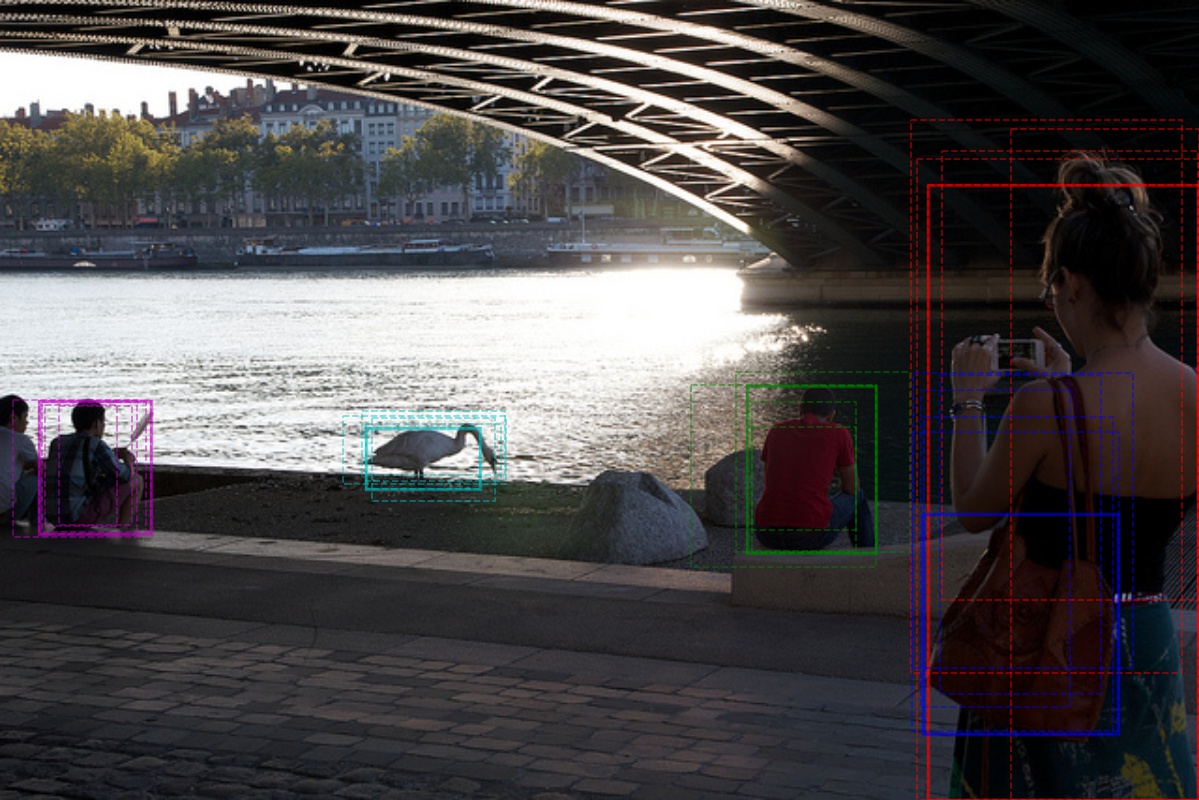}}
\subfigure{\includegraphics[height=0.3\linewidth]{5193_head34.jpg}}
\subfigure{\includegraphics[height=0.3\linewidth]{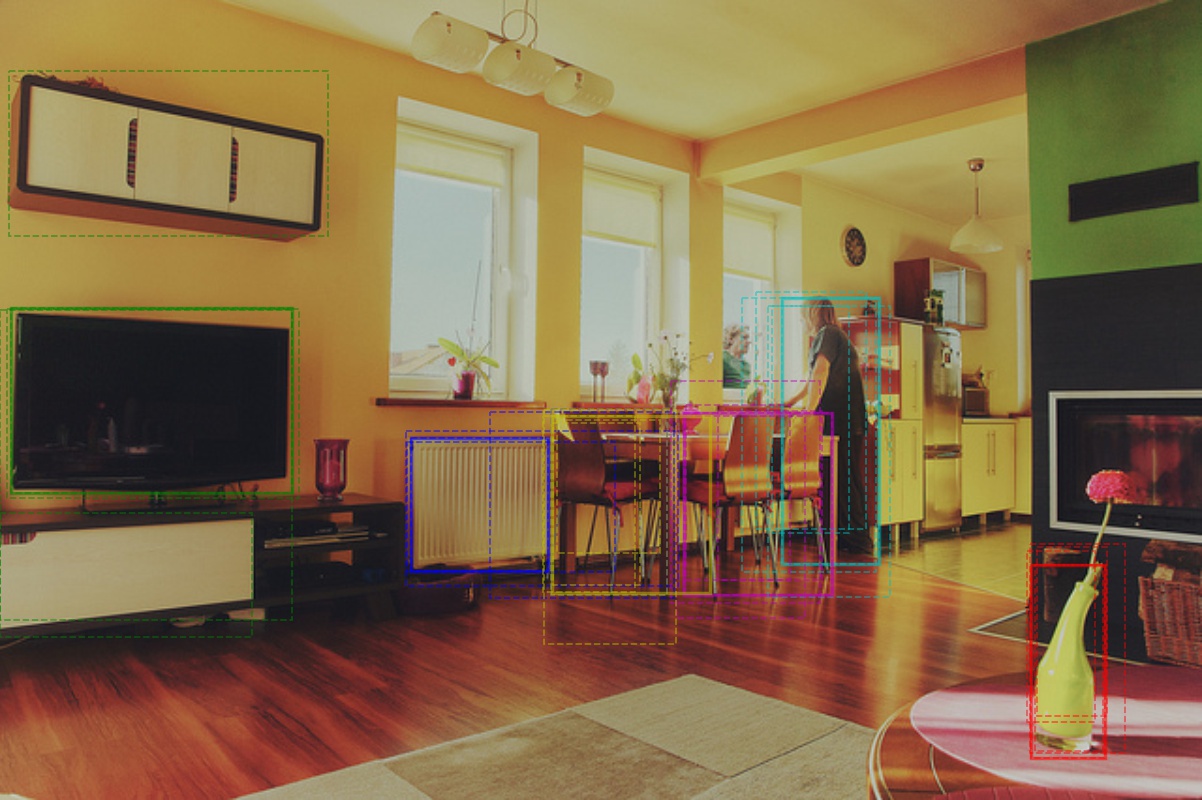}}
\subfigure{\includegraphics[height=0.3\linewidth]{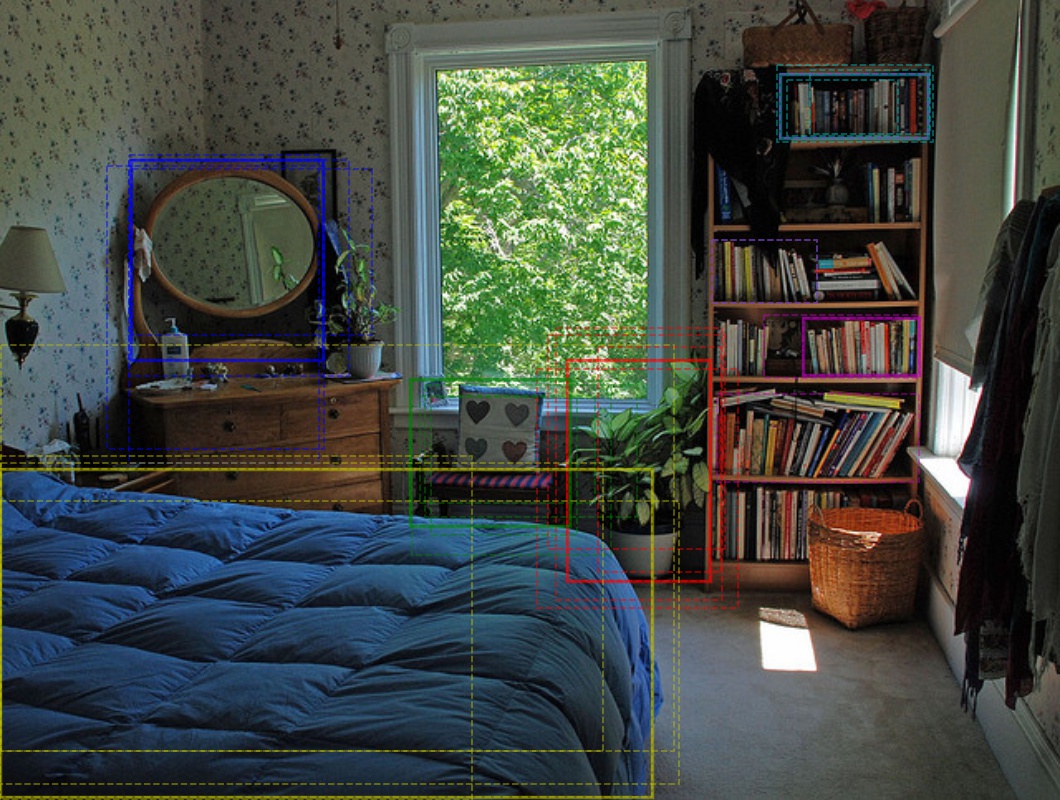}}
\subfigure{\includegraphics[height=0.3\linewidth]{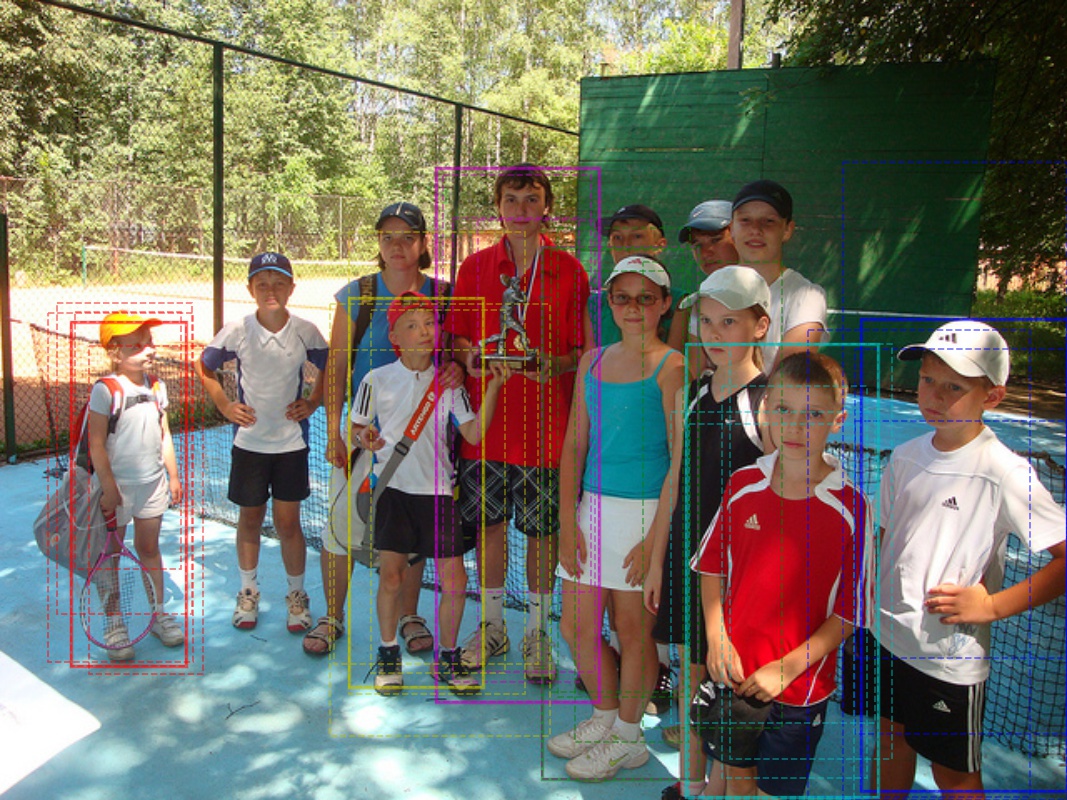}}
\subfigure{\includegraphics[height=0.3\linewidth]{2157_head34.jpg}}
\end{center}
  \caption{Qualitative analysis of TSP-RCNN on six images in the validation set. All boxes are RoI boxes predicted by RPN, where the dashed boxes are the top-5 attended boxes for the corresponding solid boxes in the same color.}
\label{fig:qualitative}
\end{figure*}

We provide a qualitative analysis of TSP-RCNN on several images in Figure \ref{fig:qualitative}. We pick one specific Transformer attention head for analysis. All boxes are RoI boxes predicted by RPN, where the dashed boxes are the top-5 attended boxes for the corresponding solid boxes in the same color. We can see that the Transformer encoder can effectively capture the RoI boxes that refer to the same instances, and hence can help to reduce the prediction redundancy.

\end{document}